\documentclass[letterpaper, 10 pt, conference]{ieeeconf}
\IEEEoverridecommandlockouts    
\usepackage{graphics}           
\usepackage{times}              
\usepackage{amsmath}            
\usepackage{amssymb}            
\usepackage{graphicx}
\usepackage{algorithm}
\usepackage[noend]{algpseudocode}
\usepackage{booktabs}
\usepackage{color}
\definecolor{instructioncolor}{rgb}{.5,.5,.5}

\usepackage[font=small]{caption}

\def\eqref#1{Eq.~(\ref{#1})}

\makeatletter
\usepackage{xspace}
\DeclareRobustCommand\onedot{\futurelet\@let@token\@onedot}
\def\@onedot{\ifx\@let@token.\else.\null\fi\xspace}

\makeatother

\usepackage{array}
\newcolumntype{L}[1]{>{\raggedright\let\newline\\\arraybackslash\hspace{0pt}}m{#1}}
\newcolumntype{C}[1]{>{\centering\let\newline\\\arraybackslash\hspace{0pt}}m{#1}}
\newcolumntype{R}[1]{>{\raggedleft\let\newline\\\arraybackslash\hspace{0pt}}m{#1}}

\usepackage{threeparttable}
\usepackage{multirow} 
\usepackage{makecell}
\usepackage{url}
\usepackage{hyperref}
\hypersetup{
    colorlinks=true,
    citecolor=blue,
    linkcolor=magenta,
    filecolor=magenta,      
    urlcolor=magenta,
}

\title{\LARGE \bf ResLPR: A LiDAR Data Restoration Network and Benchmark for \\ Robust Place Recognition Against Weather Corruptions}

\author{Wenqing\,Kuang* \and Xiongwei\,Zhao* \and Yehui\,Shen \and Congcong\,Wen \and Huimin\,Lu \and Zongtan\,Zhou \and Xieyuanli\,Chen
  \thanks{W. Kuang, Y. Shen, H. Lu, Z. Zhou, X. Chen are with 
  the College of Intelligence Science and Technology, and the National Key Laboratory of Equipment State Sensing and Smart Support, National University of Defense Technology, China.
  X. Zhao is with Harbin Institute of Technology, China. C. Wen is with New York University Abu Dhabi, UAE, and the University of Science and Technology of China, China.}
  \thanks{* indicates these authors contributed equally to this work.}  
}

\begin{document}
\maketitle
\thispagestyle{empty}
\pagestyle{empty}

\newcommand{\ourben}[0]{ResLPR}
\newcommand{\ours}[0]{ResLPRNet}
\begin{abstract}
  LiDAR-based place recognition (LPR) is a key component for autonomous driving, and its resilience to environmental corruption is critical for safety in high-stakes applications. While state-of-the-art (SOTA) LPR methods perform well in clean weather, they still struggle with weather-induced corruption commonly encountered in driving scenarios. To tackle this, we propose \ours{}, a novel LiDAR data restoration network that largely enhances LPR performance under adverse weather by restoring corrupted LiDAR scans using a wavelet transform-based network. \ours{} is efficient, lightweight and can be integrated plug-and-play with pretrained LPR models without substantial additional computational cost. Given the lack of LPR datasets under adverse weather, we introduce \ourben{}, a novel benchmark that examines SOTA LPR methods under a wide range of LiDAR distortions induced by severe snow, fog, and rain conditions. Experiments on our proposed WeatherKITTI and WeatherNCLT datasets demonstrate the resilience and notable gains achieved by using our restoration method with multiple LPR approaches in challenging weather scenarios. 
  Our code and benchmark are publicly available here: \href{https://github.com/nubot-nudt/ResLPR}{https://github.com/nubot-nudt/ResLPR}.
\end{abstract}

\section{Introduction}
\label{sec:intro}
LiDAR-based place recognition (LPR) is a key component in autonomous vehicle perception systems, with a wide range of applications, including SLAM~\cite{lcrnet, mmfvins}, global localization~\cite{chen2022overlapnet,yinsurvey} and navigation~\cite{get}. The goal of LPR is to identify the vehicle's location by matching the current LiDAR observation with location-tagged historical scans in the database~\cite{yinsurvey}. Existing LPR methods are primarily designed for good weather conditions~\cite{CVTNet,LPSNet,wen2023pyramid}, while real-world environments often involve challenging weather conditions like snow, fog, and rain. Such adverse weather can scatter, absorb, or reflect laser beams, introducing noise, interference, and even data loss in the LiDAR measurements~\cite{foglidar1,interferenc}. This degradation creates inconsistencies that influence robust retrieval of scan matching from a prebuilt database, resulting in wrong LPR results. Thus, achieving robust LPR under diverse weather conditions is a critical challenge for autonomous mobile systems.

\begin{figure}[t]
  \centering
  \includegraphics[width=0.95\linewidth]{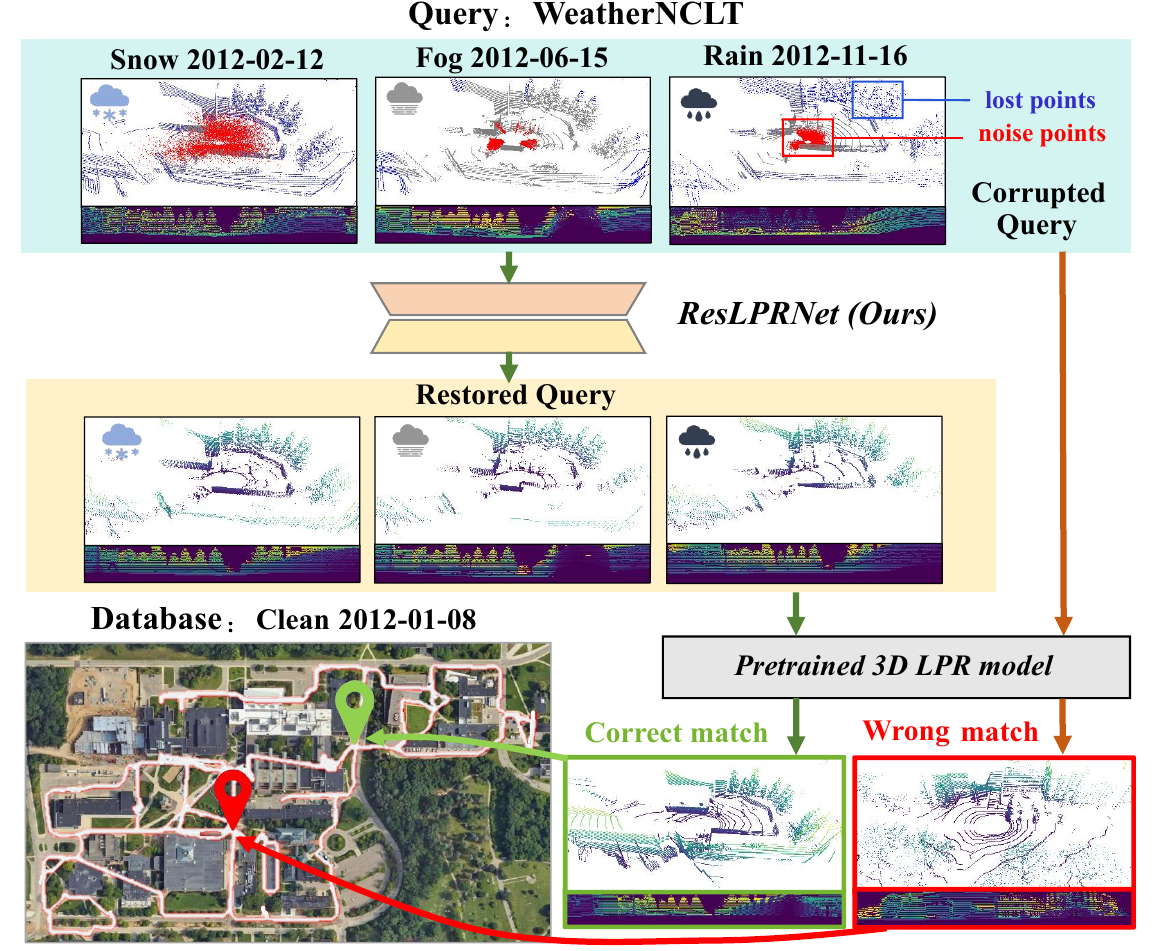}
  \caption{Our method restores LiDAR data under weather corruptions, improving the performance of a pretrained LPR model.}
  \label{ncltvis}
\end{figure}

Adverse weather conditions impact LPR tasks in two main ways: (1) introducing significant noise that distorts the spatial and intensity characteristics of point clouds; and (2) causing partial data loss due to LiDAR signal absorption. Recent studies have developed point cloud denoising methods~\cite{zhao2024triplemixer,OutDet} to reduce noise and improve downstream tasks like semantic segmentation and object detection. However, these methods focus only on denoising without addressing missing LiDAR data, thus limiting LPR accuracy. 
Moreover, current datasets~\cite{robo3d,zhao2024triplemixer} for adverse weather conditions mainly focus on semantic segmentation and object detection with dense point clouds collected over short-term operations. Benchmarks and metrics for evaluating LPR robustness under adverse weather conditions across seasons are notably lacking.

To address these challenges, we propose \ours{}, a LiDAR data restoration network that acts as a plug-and-play preprocessing module to restore weather-corrupted point clouds to near-clean clarity, thereby enhancing pretrained LPR models in adverse weather conditions, as shown in Fig.~\ref{ncltvis}. Specifically, \ours{} uses our designed WaveTransformer (WAT) block that applies wavelet transforms to isolate high-frequency noise while preserving essential structural information in the LiDAR range image. Additionally, the ContextGuide (CTG) block is designed to dynamically adapt to various weather-induced point cloud degradation patterns, guiding the restoration network to effectively remove noise and recover missing data, thereby ensuring robust LPR performance under different weather conditions.

Furthermore, we propose \ourben{}, a novel LPR benchmark with weather-corrupted point clouds. 
This benchmark includes two datasets, WeatherKITTI and WeatherNCLT, which capture three common adverse weather conditions: snow, fog, and rain, each at multiple levels of severity.
These datasets simulate LiDAR sensor measurements in real-world various weather, providing valuable data to assess LPR robustness under challenging conditions. Our benchmark assesses three LPR methods with three preprocessing approaches and introduces a new metric, stability rate, quantifying model robustness by calculating performance retention for LPR on corrupted data.

The key contributions of this work are summarized as:
\begin{itemize}
\item We propose ResLPRNet, a novel plug-and-play LiDAR restoration network that effectively removes weather-induced noise while restoring critical point cloud information, enhancing the robustness of LPR.
\item We introduce \ourben{}, a novel benchmark for evaluating LPR under adverse weather conditions, which includes two datasets, WeatherKITTI and WeatherNCLT, offering valuable resources to advance LPR research in challenging weather scenarios.
\item We evaluate three existing LPR methods enhanced by three different preprocessing techniques on our weather-corrupted benchmark and introduce a new metric, stability rate, to assess LPR robustness. 
Our benchmark and approach have been open-sourced to support the research community.
\end{itemize}

\section{Related Work}
\label{sec:related}
\subsection{LiDAR-based Place Recognition}
LPR is a classical research topic with many existing works. Early approaches focused on designing robust hand-crafted global descriptors, which can be categorized into BEV-based methods~\cite{SC}, discretization-based methods~\cite{magnusson2009appearance,cao2020season}, and point-based methods~\cite{rohling2015fast}, depending on their spatial partitioning strategies. With the advent of deep learning, neural network-based methods for global descriptor extraction have become dominant in LPR. 
Uy et al.~\cite{uy2018pointnetvlad} introduced PointNetVLAD, employing PointNet~\cite{charles2017pointnet} for local feature extraction and NetVLAD~\cite{arandjelovic2016netvlad} for global descriptor generation end-to-end. Subsequent numerous point-wise based works~\cite{liu2019lpd,LPSNet} gradually evolved on this foundation. Many efforts have also been made to explore different LiDAR data representations. For example, MinkLoc3D~\cite{minkloc3d} directly extracts local features from voxels using a 3D sparse convolutional layer. Overlap-based methods~\cite{chen2022overlapnet,ma2022overlapt} employ the overlap of range images to determine whether two point clouds are at the same location and utilize a Siamese network to estimate the overlap. CVTNet~\cite{CVTNet} combines range images and BEV images for robust LPR. 

\subsection{Robust 3D Perception Against Corruptions}
Recently, robust perception has garnered increasing attention, especially in autonomous driving tasks. Several studies~\cite{robo3d,ccr} have examined the robustness of 3D perception models trained and evaluated on public benchmarks mainly focused on 3D object detection and segmentation~\cite{kitti,nuscenes}. 
To our knowledge, no dataset or benchmark exists specifically for robustness analysis of LPR models. Some studies have attempted to enhance 3D perception robustness through data augmentation, including corrupted data during training or denoising techniques~\cite{ccr}, but these approaches have shown limited effectiveness across diverse corruption types and may impact model accuracy. Other works have introduced empirical settings and cross-density consistency training frameworks to address damage types altering point cloud density~\cite{robo3d}.
Unlike existing corruption datasets and denoising methods focusing on segmentation and detection using dense point clouds collected from short-term single operations, this paper introduces a novel benchmark and a specialized restoration network to enhance LPR performance across short-term weather changes and long-term seasonal variations.

\section{Robust Restoration Network for LPR}
\label{sec:main}

\begin{figure*}[t]
  \centering
  \captionsetup{aboveskip=2pt, belowskip=0pt}
  \includegraphics[width=\linewidth]{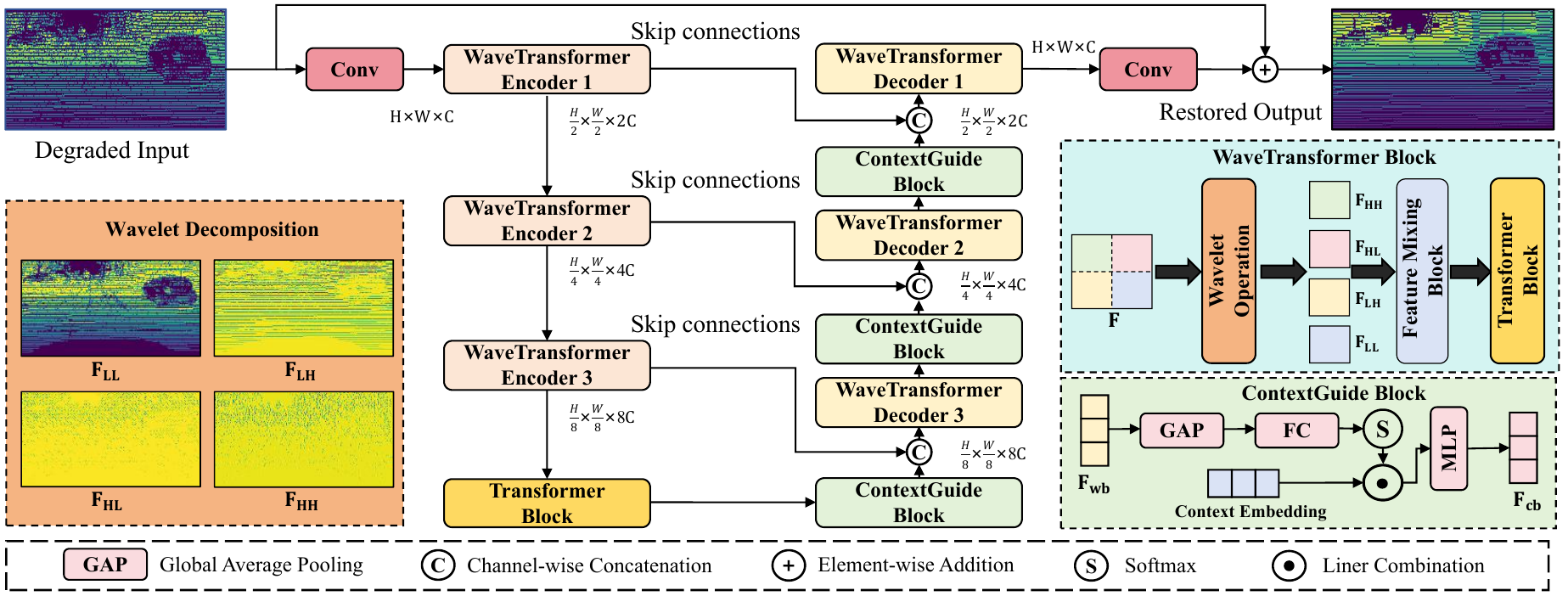}
   \caption{Pipeline of \ours{}. It adopts a hierarchical encoder-decoder architecture to restore clean LiDAR data from noisy weather-induced inputs. The three-layer encoder and decoder employ similar WaveTransformer blocks, with the encoder performing wavelet decomposition and the decoder executing wavelet reconstruction. An additional Transformer block is applied to the bottleneck feature to enhance its spatial correlation. ContextGuide blocks in each decoder layer enable the network to adapt to different degradation.}
   \label{restore_ov}
   \vspace{-0.5cm}
\end{figure*}

\subsection{Overall Pipeline}
Fig.~\ref{restore_ov} shows the pipeline of our \ours{}, which leverages the multi-scale analysis properties of wavelet transforms and contextual information from various weather-induced degradation patterns to achieve robust LiDAR data denoising and reconstruction.

Specifically, given a corrupted point cloud $\mathcal{P}_{c}\in\mathbb{R}^{N\times4}$, we project it into a 2D range image $\mathcal{I}_{c}\in\mathbb{R}^{H\times W\times C}$ using spherical projection~\cite{chen2022overlapnet,ma2022overlapt}, where $H \times W$ represents the spatial dimensions of the projection and $C$ denotes the number of channels. Each channel encodes features, including the normalized distance $d$ and intensity $i$ values of each point. \ours{} first embeds the input LiDAR range image $\mathcal{I}_{c}$ to obtain initial features $\mathbf{F}_0\in\mathbb{R}^{H\times W\times C}$ by applying a convolution operation. $\mathbf{F}_0$ is then passed through a hierarchical encoder-decoder framework designed to progressively enrich and refine the feature representations. The encoder is structured into three layers, each using WaveTransformer (WAT) blocks that integrate wavelet decomposition, feature mixing, and a transformer block. As features pass through these layers, spatial resolution reduces while channel count increases, yielding a compact latent representation $\mathbf{F}_\text{d}\in\mathbb{R}^{\frac{H}{8} \times \frac{W}{8} \times 8C}$. An additional transformer block is applied on the bottleneck feature to capture the spatial correlation thus enhancing the feature representation ability.
Then, a three-layer decoder comprising WaveTransformer blocks performs wavelet reconstruction, where restored features $\mathbf{F}_\text{r}\in\mathbb{R}^{{H} \times {W} \times C}$ are gradually reconstructed from $\mathbf{F}_\text{d}$ to recover spatial details. To enhance the restoration against diverse weather conditions, we introduce ContextGuide (CTG) blocks at each decoder layer. Skip connections between encoders and decoders preserve crucial spatial details, ensuring high-fidelity restoration. Detailed descriptions of the WaveTransformer and ContextGuide block designs are as follows.

\subsection{WaveTransformer Block}
Restoring LiDAR data from 2D range images requires pixel-level correspondence between degraded and clean images, but resampling often results in detail loss, limiting the network capacity to capture important features. To address this, we propose to use wavelet transforms instead of conventional up- and down-sampling, allowing multi-scale analysis without compromising spatial resolution. Building on this idea, as shown in Fig. \ref{restore_ov}, we design the WaveTransformer block to efficiently capture global and local information across encoders and decoders. Each block extracts multi-scale features via wavelet decomposition or reconstruction, followed by feature mixing for detailed spatial and channel information. A final transformer block enhances global dependencies, enhancing feature representation across scales. Given an input feature embedding $\mathbf{F} \in \mathbb{R}^{H \times W \times C}$, where $\mathbf{F} = \mathbf{F}_0$ for the first block, and subsequent blocks $\mathbf{F}$ represents the output feature from the previous block, the discrete wavelet transform (DWT) decomposes it into four wavelet sub-bands:
\begin{equation}
\mathbf{F}_\text{LL}, \mathbf{F}_\text{LH}, \mathbf{F}_\text{HL}, \mathbf{F}_\text{HH} = \text{DWT}(\mathbf{F}),
\end{equation}
where $\mathbf{F}_\text{LL}, \mathbf{F}_\text{LH}, \mathbf{F}_\text{HL}, \mathbf{F}_\text{HH} \in\mathbb{R}^{\frac{H}{2}\times \frac{H}{2}\times C}$ are four sub-images with different frequencies. 
As shown in Fig.~\ref{restore_ov}, the low-frequency components $\mathbf{F}_\text{LL}$ retain the overall structural features of the range image, while the high-frequency horizontal $\mathbf{F}_\text{LH}$, vertical $\mathbf{F}_\text{HL}$, and diagonal $\mathbf{F}_\text{HH}$ components effectively isolate noise information. 
During the encoder’s wavelet decomposition stage, we directly concatenate these four components to form the shallow feature $\mathbf{F}_\text{ws}\in\mathbb{R}^{\frac{H}{2} \times \frac{W}{2} \times 4C}$. In the decoder’s wavelet reconstruction, we use transposed convolution for upsampling to obtain 
$\mathbf{F}_\text{w}\in\mathbb{R}^{{H} \times {W} \times C}$:
\begin{equation}
\mathbf{F}_\text{w} = \text{ConvTrans}(\text{Concat}(\mathbf{F}_\text{LL}, \mathbf{F}_\text{LH}, \mathbf{F}_\text{HL}, \mathbf{F}_\text{HH})),
\end{equation}

\begin{figure}[t]
  \captionsetup{aboveskip=2pt, belowskip=0pt}\centerline{\includegraphics[width=\linewidth]{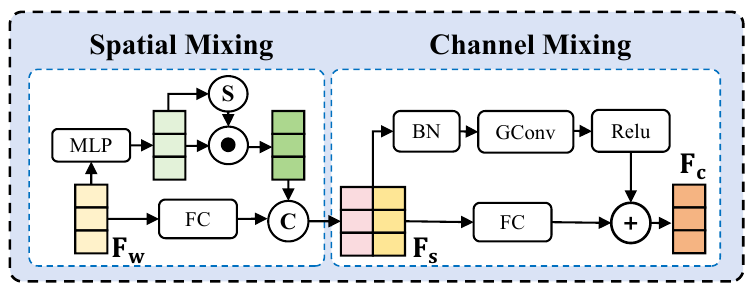}}
	\caption{Details of the Feature Mixing Block, which includes two operations: spatial mixing and channel mixing. BN denotes batch normalization,  GConv represents grouped convolution and FC is fully connected layer. Other symbol definitions are the same as in Fig. \ref{restore_ov}.}
	\label{fmb}
\vspace{-0.5cm}
\end{figure}

After obtaining the features $\mathbf{F}_\text{w}$ from the wavelet operation, the feature mixing block applies spatial and channel mixing to capture detailed spatial and contextual features, as shown in Fig. \ref{fmb}. The spatial mixing operation uses a self-attention mechanism to extract contextual spatial features $\mathbf{F}_\text{s}$. Next, the channel mixing operation employs a grouped convolution to capture inter-channel dependencies and facilitate efficient channel-wise feature learning, resulting in a refined feature representation $\mathbf{F}_\text{c}$.

Finally, we apply a transformer block to fuse the original features $\mathbf{F}_\text{w}$ with the enhanced features $\mathbf{F}_\text{c}$ obtained from the feature mixing block, capturing long-range dependencies and enriching the representation with detailed local and global context:
\begin{equation}
\operatorname{Attention}(\mathbf{Q}, \mathbf{K}, \mathbf{V})=\operatorname{softmax}\left(\frac{\mathbf{Q} \mathbf{K}^\top}{\sqrt{d_{k}}}\right) \mathbf{V},
\end{equation}
\begin{equation}
\mathbf{F}_{\text{att}} = \text{Norm}(\mathbf{F}_\text{w} + \text{Attention}(\mathbf{Q}, \mathbf{K}, \mathbf{V})),
\end{equation}
\begin{equation}
\mathbf{F}_\text{wb} = \mathbf{F}_{\text{att}} +\text{FFN}(\mathbf{F}_{\text{att}}),
\end{equation}
where $\mathbf{Q} = \mathbf{F}_\text{w}, \mathbf{K} = \mathbf{F}_\text{c}, \mathbf{V} = \mathbf{F}_\text{c}$, $d_{k}$ is the dimensionality of the key. The attention-enhanced features $\mathbf{F}_{\text{att}}$ are then passed through a feed-forward network (FFN) to produce the final output feature $\mathbf{F}_\text{wb}$ of the WaveTransformer block.

\subsection{ContextGuide Block}
LiDAR data exhibit varying degradation patterns under different weather conditions. The ContextGuide block enhances the model's adaptability by dynamically adjusting feature representations to handle diverse degradation patterns. The structure of the ContextGuide block is shown in Fig.~\ref{restore_ov}. We introduce a learnable parameter, Context Embedding ($\mathbf{CE}$), which interacts with the incoming features $\mathbf{F}_\text{wb}$ to embed degradation information, thereby adapting to the features of different degradation patterns.
Global average pooling (GAP) is applied to the input feature $\mathbf{F}_\text{wb}$ to extract global information. The GAP output is passed through a fully connected layer and softmax to generate weights $\mathbf{W}_\text{wb}$, which are then multiplied with $\mathbf{CE}$ and summed. The resulting feature undergoes convolution to enhance spatial coherence:
\begin{equation}
\mathbf{W}_\text{wb} = \text{Softmax}(\text{FC}(\text{GAP}(\mathbf{F}_\text{wb})),
\end{equation}
\begin{equation}
\mathbf{F}_\text{cb} = \text{MLP}(\sum_{} \mathbf{W}_\text{wb} \cdot \mathbf{CE}),
\end{equation}
where $\mathbf{F}_\text{cb}$ contains input features with different degradation information. By leveraging contextual information, the ContextGuide block effectively enhances the model's robustness. 

\subsection{Loss Function}
To train our \ours{} and ensure it effectively target on LiDAR data restoration, we define the loss function $L$ as:
\begin{equation}
L = \frac{1}{N} \sum_{j=1}^{N} \left( |d_j - \hat{d}_j| + |i_j - \hat{i}_j| \right),
\end{equation}
where $d_j$ and $i_j$ represent the original distance and intensity measurements in the clean LiDAR range image, $\hat{d}_j$ and $\hat{i}_j$ are the model's predicted values, and $N$ is the number of elements. This loss function penalizes the absolute differences between the predicted and original measurements, promoting robustness to outliers and stable convergence. By minimizing this loss, the model learns to accurately restore the original measurements, including range and intensity, in degraded LiDAR data, resulting in smooth and reliable point cloud restoration.

\section{Weather-Induced LPR Benchmark}

We establish a novel benchmark specifically designed for evaluating LPR resilience in adverse weather. Our benchmark categorizes three types of weather-induced corruption commonly encountered in real-world deployments: snow, fog, and rain, as shown in Fig.~\ref{ncltvis} and Fig.~\ref{query_vis}. This section provides a detailed overview of the generation of each corruption type and the benchmark configuration. 
More visualizations are provided in our online supplementary\footnote{\url{https://github.com/nubot-nudt/ResLPR/blob/main/supplementary.pdf}}.

\subsection{Weather Corruptions Generation}
\label{weather_generation}

Given a LiDAR point \(\mathbf{p} \in \mathbb{R}^4 \), where \( \mathbf{p} \)  has coordinates \( (x, y, z) \) and intensity \( i \), we apply a transformation \( \hat{\mathbf{p}} = T(\mathbf{p}) \) to generate a corrupted point \( \hat{\mathbf{p}} \). 
The underlying principles for each corruption are detailed as follows.

\begin{table}[t]
\scriptsize
\captionsetup{aboveskip=2pt, belowskip=0pt}
\renewcommand\arraystretch{0.9}
\setlength{\tabcolsep}{3pt}
\centering
\caption{Dataset Division for WeatherKITTI and WeatherNCLT}
\small
\label{tab:dataset}
\renewcommand{\arraystretch}{1.0} 
\small
\begin{threeparttable}
\begin{tabular}{cccc}
\toprule
\multicolumn{2}{c}{} & \textbf{WeatherKITTI} & \textbf{WeatherNCLT} \\
\midrule
\addlinespace[2pt]

\multicolumn{2}{c}{\textbf{LiDAR beams}} & 64 & 32 \\
\midrule
\multicolumn{2}{c}{\textbf{Frame counts}} & 156K & 1,270K \\
\midrule
\multicolumn{2}{c}{\multirow{3}{*}{\centering \vspace{2.5em}\textbf{Train}}} & \makecell{S03 to S10 \\ F03 to F10 \\ R03 to R10} & \makecell{S2012-01-22 \\ F2012-08-04 \\ R2012-11-04} \\
\midrule
\multirow{3}{*}{\textbf{Test}} & \multirow{2}{*}{\centering\textbf{Query}} & S00, F00, & S2012-02-12, F2012-06-15, \\ & & R00 & R2012-11-16 \\[1pt]
\cline{2-4}
\rule{0pt}{10pt}& \textbf{Database} & C00  & C2012-01-08 \\
\bottomrule
\end{tabular}
\begin{tablenotes}
\footnotesize
\centering
\item S: snow, F: fog, R: rain, C:  original clean sequence.
\end{tablenotes}
\end{threeparttable}
\vspace{-0.5cm}
\end{table}

\subsubsection{Fog-Induced Corruption}
LiDAR sensors emit laser pulses to measure distances. In foggy conditions, water particles induce backscattering and attenuation, leading to pulse reflections~\cite{foglidar1,fogsim} that alter the point cloud's coordinates and intensities. This effect generates densely clustered noisy points near the LiDAR sensor. Following~\cite{fogsim}, we simulate fog-attenuated point $\mathbf{p_{f}}$ as:
\begin{equation}
{{\mathbf{p_{{f}}}}}=\left\{\begin{array}{ll}
\left(s\cdot(x, y, z), {i_{\text {soft }}}\right), & \text { if } {i_{\text {soft }}}>{i_{\text {hard }}}, \\
\left(x, y, z, {i_{\text {hard }}}\right), & \text { else, } 
\end{array}\right.
\end{equation}
where $s$ is a scaling factor that depends on the range of the reflective object. $i_{\text{soft }}$ and $i_{\text{hard }}$ represent the received responses for the soft target and hard target, defined as:
\begin{equation}
i_{\text{hard }}={i} \cdot \exp ({-2 \alpha R_{0}}),\quad
i_{\text {soft}}={i} \cdot {R_{0}^{2}} \cdot {\beta} \cdot {i}_{t},
\end{equation}
where $R_{0}$ denotes the range of the hard target, $\alpha$ is the attenuation coefficient, $\beta$ represents the ratio of the noise factor to the target's differential reflectivity. ${i}_{t}$ is the maximum response of the soft-target term. As $\beta$ increases, the fog density rises, resulting in more fog particles in the point cloud.

\subsubsection{Snow-Induced Corruption}

In snowy conditions, each LiDAR beam interacts with airborne particles, affecting reflection angles and causing occlusions~\cite{lidarsnow,snowsim}. Following~\cite{snowsim}, we simulate snow particles in 2D and modify each LiDAR beam measurement based on the snowfall rate $r_s$. The snow-attenuated point $\mathbf{p_{s}}$ is calculated as follows:
\begin{equation}
i_{\text{snow}} = T_{R}+i_{\max } f_{s}\left|f_{o}-\left(1-\frac{R^{*}}{R_{\max }}\right)\right|^{2}, \\
\end{equation}
\begin{equation}
{{\mathbf{p_{\text{s}}}}}= \left(\frac{r_s}{\gamma} \cdot (x, y, z), i_{\text{snow}}\right),
\end{equation}
where \( R^{*} \) and \( R_{\max} \) are the distances to the maximum power \( T_{R} \) and the farthest detectable object, respectively. \( f_{s} \) and \( f_{o} \) denote the LiDAR sensor's focal slope and offset, while \( \gamma\) is the target’s differential reflectivity coefficient. Higher snowfall rates \( r_s \), lead to greater spatial variation and denser particle accumulations in the point cloud.

\subsubsection{Rain-Induced Corruption}
In rainy weather, LiDAR beams interact with raindrops, causing reflections and partial occlusions, which alter beam intensity and add noise to the point cloud. We use a physics-based simulation model~\cite{snowsim} with a Monte Carlo approach to simulate rain conditions. Similar to snow corruption, the number of rain particles is adjusted based on a specified rain rate $r_r$.

\subsection{Benchmark Configuration}
\label{configuration}

\subsubsection{Corruption Datasets}

To evaluate the corruption robustness of LPR methods, we constructed the WeatherKITTI and WeatherNCLT datasets based on the widely used LPR datasets KITTI~\cite{kitti} and NCLT~\cite{nclt}, utilizing the aforementioned simulation methods. Each dataset includes three types of weather-induced corruption, with each type exhibiting three severity levels. This resulted in a total of 156,951 and 1,270,314 annotated LiDAR scans, as detailed in Table~\ref{tab:dataset}. Specific simulation parameters can be found in the supplementary material.

\subsubsection{WeatherKITTI}

It aims to test short-term loop closure detection robustness under weather changes. Following~\cite{CVTNet}, we use sequences 03 to 10 for training and 00 to evaluate the LPR performance. We use the clean sequence 00 as the database, while the query scans are with various types and severity levels of the corrupted sequence 00. To train our ResLPRNet and other preprocessing methods, we use corrupted-clean pairs from sequences 03 to 10.

\subsubsection{WeatherNCLT}

It aims to evaluate long-term LPR under adverse weather over an extended period. To replicate the original weather conditions of the NCLT dataset, we simulated snow for sequence 2012-02-12, fog for 2012-06-15, and rain for 2012-11-16 as queries, with the clean sequence from 2012-01-08 serving as the database, following~\cite{CVTNet}. Additionally, we generated corrupted 2012-01-22, 2012-08-04, and 2012-11-04 sequences to train preprocessing networks.

\subsubsection{Evaluated LPR Methods}
We benchmark three different types of LPR methods: SC~\cite{SC}, CVTNet~\cite{CVTNet}, and LPSNet~\cite{LPSNet}. SC~\cite{SC} is a traditional method that identifies places by comparing point clouds based on their distribution characteristics in 2D space. CVTNet~\cite{CVTNet}, a learning-based approach using multi-view LiDAR images, achieves strong performance on clean LiDAR datasets. LPSNet~\cite{LPSNet} is also a learning-based method that directly processes 3D point clouds demonstrating impressive LPR results on clean data. In our benchmarking, we apply the pretrained models of CVTNet and LPSNet, trained with the protocol from~\cite{CVTNet} using the original dataset, to simulate the impact of unpredictable weather corruption in real-world applications.

\subsubsection{Boosting Corruption Robustness}

Since no existing methods are designed for preprocessing corrupted LiDAR data for LPR under severe weather, our benchmark compares \ours{} with two baseline denoising methods: WeatherNet~\cite{weathernet} and TripleMixer~\cite{zhao2024triplemixer}. WeatherNet filters weather effects by learning the differences between noise points and surrounding pixels in LiDAR range images. 
TripleMixer comprises three mixer layers to capture multi-view, multi-scale features, effectively removing noise.

\section{Experiments}
\label{sec:exp}
\subsection{Implement Details}

For \ours{}, input range images $\mathcal{I}_{c}$ in WeatherKITTI are sized $64 \times 1920 \times 2$ and in  WeatherNCLT $32 \times 1440 \times 2$, aligning with the LiDAR beam numbers and the average point counts per LiDAR scan in each dataset. To conduct point cloud restoration on  WeatherKITTI and  WeatherNCLT, we trained the network using the corrupted-clean LiDAR scan pairs from both datasets provided in our benchmark without extra manual labeling. We utilized the Adam optimizer with a learning rate of $1e{-}4$ and trained the network with 120 epochs. During training, a cropping patch of size $32 \times 480$ was used as input, and random horizontal and vertical flips were applied to the input image to augment the training data.  

\begin{table*}[h]
\centering
\footnotesize
\captionsetup{aboveskip=2pt, belowskip=0pt}
\renewcommand\arraystretch{0.9}
\setlength{\tabcolsep}{3.8pt}
\caption{\centering{The benchmarking results of LPR methods under three preprocessing methods on WeatherKITTI.}}
\begin{tabular}{lcccccccccccccc}
\toprule
\multicolumn{2}{c}{\textbf{Method}} &\multirow{2}{*}{\textbf{$\text{mSR}_{l}$↑}}  & \multicolumn{4}{c}{\textbf{KITTI Snow}} & \multicolumn{4}{c}{\textbf{KITTI Fog}} & \multicolumn{4}{c}{\textbf{KITTI Rain}} \\

\cmidrule(lr){1-2} \cmidrule(lr){4-7} \cmidrule(lr){8-11} \cmidrule(lr){12-15}
\textbf{Preprocessing}  &\textbf{LPR}  & & \textbf{AUC↑} & \textbf{F1↑} & \textbf{R@1↑} & \textbf{R@5↑} 
&\textbf{AUC↑} & \textbf{F1↑} & \textbf{R@1↑} & \textbf{R@5↑} 
& \textbf{AUC↑} & \textbf{F1↑} & \textbf{R@1↑} & \textbf{R@5↑} \\
\midrule
\multirow{3}{*}{\textbf{Corruption}} &SC \cite{SC}&0.310 &0.140	&0.180	&0.161	&0.176&0.346	&0.383	&0.335	&0.357	&0.193	&0.231	&0.200	&0.230
\\
&CVTNet \cite{CVTNet}		&0.192   &0.024	&0.051	&0.079 &0.139	&0.113	&0.138	&0.185	&0.276 &0.089	&0.125	&0.187	&0.308\\
&LPSNet \cite{LPSNet}	&0.360	&0.033	&0.066	&0.123	&0.203 &0.126	&0.181	&0.326	&0.441	&0.185	&0.255	&0.478	&0.653\\
\midrule
\multirow{3}{*}{\textbf{WeatherNet}\cite{weathernet}} &SC \cite{SC} &0.424 	&0.249	&0.300	&0.226	&0.237	&0.314 &0.362	&0.273	&0.292 &0.445	&0.489	&0.410	&0.422
\\
&CVTNet \cite{CVTNet}		&0.112 	&0.005	&0.012	&0.017 &0.042	&0.027	&0.049	&0.061	&0.103 &0.083	&0.124	&0.185	&0.299\\
&LPSNet \cite{LPSNet}	&0.418	&0.092	&0.142	&0.252	&0.367 &0.097	&0.146	&0.251	&0.366	&0.321	&0.337	&0.573	&0.707\\
\midrule 
\multirow{3}{*}{\textbf{TripleMixer}\cite{zhao2024triplemixer}} &SC \cite{SC} &0.496 	&0.168	&0.203	&0.180	&0.201	&0.627	&0.650	&0.597	&0.618 &0.457	&0.508	&0.413	&0.427
\\
&CVTNet	\cite{CVTNet}	& 0.226 	&0.046	&0.082	&0.133 &0.216 &0.145	&0.174	&0.218	&0.290	&0.099	&0.135	&0.203	&0.318\\
&LPSNet \cite{LPSNet}	&0.551	&0.233	&0.274	&0.466	&0.617	&0.182	&0.232	&0.436	&0.564 &0.330	&0.352	&0.582	&0.725\\
\midrule 
\multirow{3}{*}{\textbf{ResLPR(Ours)}} &SC \cite{SC} &\textbf{0.949}  &\textbf{0.804}	&\textbf{0.768}	&\textbf{0.803}	&\textbf{0.840}	&\textbf{0.793}	&\textbf{0.759} &\textbf{0.798} &\textbf{0.832} &\textbf{0.815}	&\textbf{0.799}	&\textbf{0.806}	&\textbf{0.839}
\\
&CVTNet \cite{CVTNet}		& \textbf{0.794} 	&\textbf{0.612}	&\textbf{0.534}	&\textbf{0.793}	&\textbf{0.865}	&\textbf{0.361}	&\textbf{0.340}	&\textbf{0.681}	&\textbf{0.820} &\textbf{0.760}	&\textbf{0.692}	&\textbf{0.843}	&\textbf{0.883}
\\
&LPSNet \cite{LPSNet}	&\textbf{0.915}	&\textbf{0.524}	&\textbf{0.560}	&\textbf{0.833}	&\textbf{0.884}	&\textbf{0.490}	&\textbf{0.564}	&\textbf{0.820}	&\textbf{0.868} &\textbf{0.549}	&\textbf{0.577}	&\textbf{0.838}	&\textbf{0.880}\\
\bottomrule
\label{ WeatherKITTI}
\end{tabular}
\vspace{-0.5cm}
\end{table*}

\begin{table*}[h]
\centering
\footnotesize
\captionsetup{aboveskip=2pt, belowskip=0pt}
\renewcommand\arraystretch{0.9}
\setlength{\tabcolsep}{6.6pt}
\caption{\centering{The benchmarking results of LPR methods under three preprocessing methods on WeatherNCLT.}}
\begin{tabular}{lccccccccccc}
\toprule
\multicolumn{2}{c}{\textbf{Method}} &\multirow{2}{*}{\textbf{$\text{mSR}_{p}$↑}}  & \multicolumn{3}{c}{\textbf{NCLT Snow}} & \multicolumn{3}{c}{\textbf{NCLT Fog}} & \multicolumn{3}{c}{\textbf{NCLT Rain}} \\

\cmidrule(lr){1-2} \cmidrule(lr){4-6} \cmidrule(lr){7-9} \cmidrule(lr){10-12}
\textbf{Preprocessing}  &\textbf{LPR} & & \textbf{R@1↑} & \textbf{R@5↑} & \textbf{R@20↑}
& \textbf{R@1↑} & \textbf{R@5↑} & \textbf{R@20↑}
& \textbf{R@1↑} & \textbf{R@5↑} & \textbf{R@20↑} \\
\midrule
\multirow{3}{*}{\textbf{Corruption}} &SC \cite{SC}&0.148	&0.124	&0.138	&0.154	&0.069	&0.075	&0.077 &0.083	&0.100	&0.122

\\
&CVTNet	\cite{CVTNet}	&0  	&-	&-	&0.004	&-	&-	&- &-	&-	&0.004
\\
&LPSNet \cite{LPSNet}	&0.187		&0.161	&0.239	&0.318	&0.037	&0.048	&0.054 &0.067	&0.088	&0.106\\
\midrule
\multirow{3}{*}{\textbf{WeatherNet}\cite{weathernet}} &SC \cite{SC} &0.225 	&0.329	&0.345	&0.364	&0.071	&0.077	&0.081 &0.088	&0.106	&0.122
\\
&CVTNet	\cite{CVTNet}	&0.380  	&0.750	&0.822	&0.877	&0.123	&0.156	&0.186 &0.055 &0.073	&0.104	\\
&LPSNet \cite{LPSNet}	&0.205	&0.195	&0.293	&0.380 &0.040	&0.049	&0.057	&0.064	&0.084	&0.105	\\
\midrule 
\multirow{3}{*}{\textbf{TripleMixer}\cite{zhao2024triplemixer}} &SC \cite{SC}&0.159	&0.141	&0.156	&0.170	&0.071	&0.077	&0.081 &0.091	&0.108	&0.124
\\
&CVTNet	\cite{CVTNet}	&0.419  	&0.813 &0.877	&0.922	&0.123	&0.157	&0.187 &0.083 &0.111	&0.136	\\
&LPSNet \cite{LPSNet}	&0.224		&0.223	&0.326	&0.421	&0.038	&0.049	&0.061 &0.074	&0.093	&0.112	\\
\midrule 
\multirow{3}{*}{\textbf{ResLPR(Ours)}} &SC \cite{SC} &\textbf{0.705} 	&\textbf{0.597}	&\textbf{0.620}	&\textbf{0.637}&\textbf{0.580}	&\textbf{0.598}	&\textbf{0.607}	&\textbf{0.322}	&\textbf{0.357}	&\textbf{0.386}
\\
&CVTNet \cite{CVTNet}		&\textbf{0.920}  	&\textbf{0.898}	&\textbf{0.935}	&\textbf{0.961}	&\textbf{0.785}	&\textbf{0.852}	&\textbf{0.914} &\textbf{0.604}	&\textbf{0.709}	&\textbf{0.805}
\\
&LPSNet \cite{LPSNet}	&\textbf{0.921}	&\textbf{0.662}	&\textbf{0.765}	&\textbf{0.812}	&\textbf{0.517}	&\textbf{0.685}	&\textbf{0.770} &\textbf{0.321}	&\textbf{0.457}	&\textbf{0.555}
\\
\bottomrule
\label{ WeatherNCLT}
\end{tabular}
\vspace{-0.6cm}
\end{table*}

\subsection{Evaluation Metrics}
We employ different metrics to evaluate the performance of the LPR methods. In addition to the classic LPR metrics outlined in~\cite{CVTNet}, including $\text{Recall@N}$, $\text{AUC}$, and $\text{F1}$, we introduce a novel metric, the mean stability rate ($\text{mSR}$), as a relative measure for evaluating the robustness of LPR methods in maintaining performance on corrupted datasets. On WeatherKITTI, we calculate $\text{mSR}_{l}$ as: 
\begin{equation}
\text{Alp}=\sum(\text{AUC},\text{F1},\text{R@1},\text{R@5}),
\end{equation}
\begin{equation}
{\text{SR}}_{li}=\frac{\sum_{s=1}^{3} \mathit{\text{Alp}}_{i, s}}{3 \times {\text{Alp}}_{\text {clean}}}, \quad {\text{mSR}}_{l}=\frac{1}{N} \sum_{i=1}^{N} {\text{SR}}_{li},
\label{eq:mSR}
\end{equation}
where $\text{{Alp}}_{i, s}$ denotes the overall performance on corruption type $i$ at severity level $s$, and $\text{{Alp}}_{\text {clean}}$ represents the performance of the LPR model in the clean dataset. $N$ = 3 is the total number of corruption types. The robustness indicator on  WeatherNCLT is $\text{mSR}_{p}$, and its calculation method remains the same except that $\text{Alp} = \sum(\text{R@1}, \text{R@5}, \text{R@20})$.

\subsection{Performance Analysis}
This section presents LPR robustness evaluation results on our benchmark. The performance of SC~\cite{SC}, CVTNet~\cite{CVTNet}, and LPSNet~\cite{LPSNet} on original clean datasets are referred to~\cite{CVTNet} or supplementary material due to page limitations.

\subsubsection{Results on WeatherKITTI}
The first experiment evaluates preprocessing and LPR methods with the  WeatherKITTI 64-beam LiDAR dataset under typical driving scenarios. We present the performance restoration results for three LPR approaches using three preprocessing methods on corrupted  WeatherKITTI data in Table~\ref{ WeatherKITTI}. 
Note that we report results for moderate-level corruption, with full results for other severity levels provided in the supplementary materials due to page limitations.
As observed, both learning-based and hand-crafted feature-based LPR methods show substantial performance degradation under adverse weather conditions. However, after applying different preprocessing methods, all LPR methods obtain consistent performance improvements to varying degrees, validating the effectiveness of denoising and restoration techniques for enhancing LPR in challenging weather. 
Notably, our \ours{} demonstrates the largest performance gains across all three LPR methods, outperforming the other two denoising approaches across all weather-induced corruptions and evaluation metrics. 
For example, in the KITTI-Snow cases, the $\text{AUC}$ of SC~\cite{SC} drops to just 0.14. However, after restoring the corrupted LiDAR data by our \ours{}, the $\text{AUC}$ improves significantly to 0.804, nearly matching the performance on clean data provided in the supplementary. Examining the robustness $\text{mSR}_{l}$, we find that \ours{}-enhanced LPR algorithm achieves a relatively high robustness value and greatly surpasses those of the other two denoising methods. 
Overall, the experimental results indicate that our restoration method helps maintain strong LPR robustness under different adverse weather conditions. 

\begin{figure}[t]
  \captionsetup{aboveskip=2pt, belowskip=0pt}\centerline{\includegraphics[width=\linewidth]{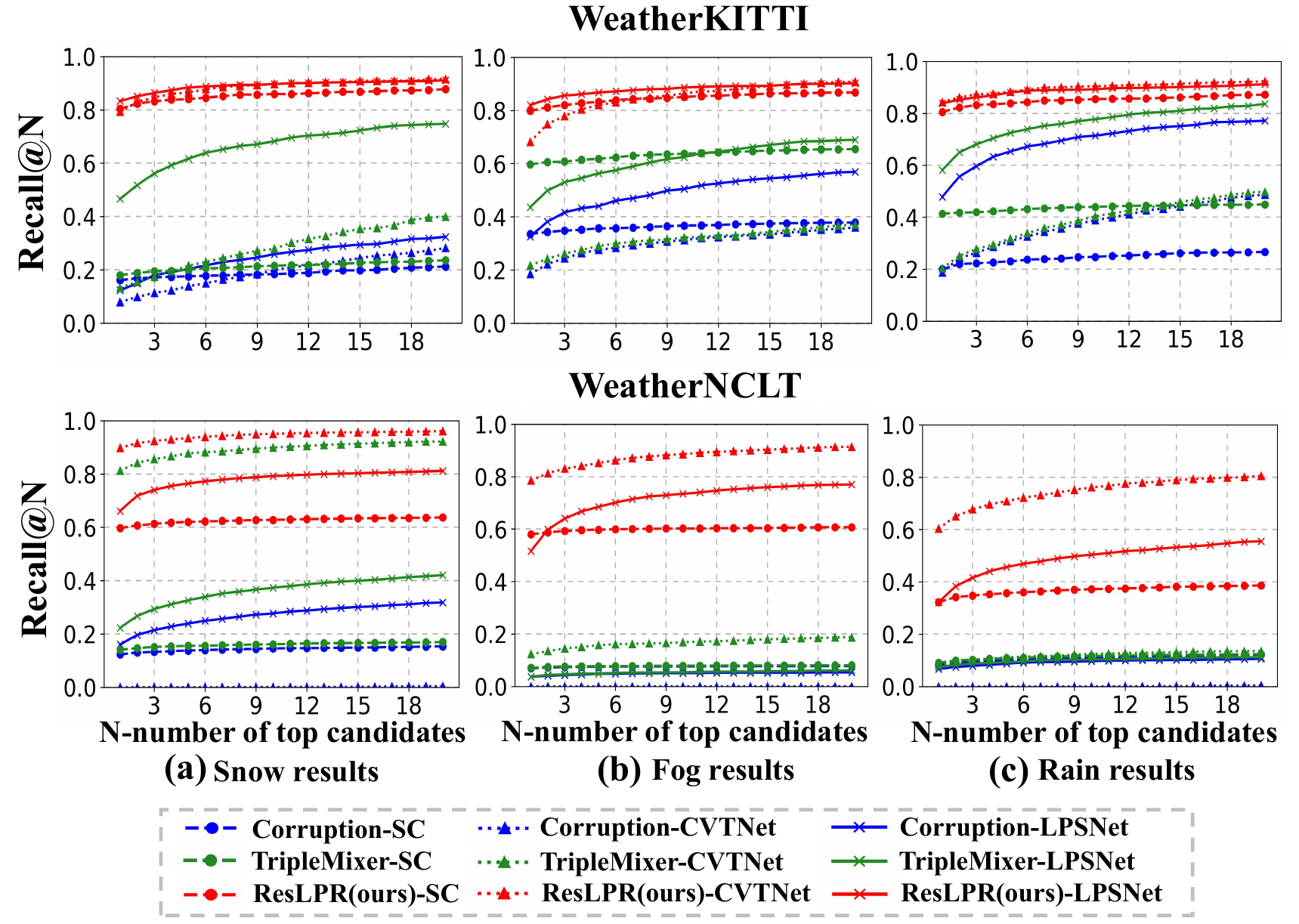}}
	\caption{Recall@N curves on the WeatherKITTI and WeatherNCLT datasets. (a) Snow results (b) Fog results (c) Rain results}
	\label{recall@N}
\vspace{-0.5cm}
\end{figure}

\begin{figure*}[ht]
    \captionsetup{aboveskip=2pt, belowskip=0pt}
    \centerline{\includegraphics[width=\linewidth]{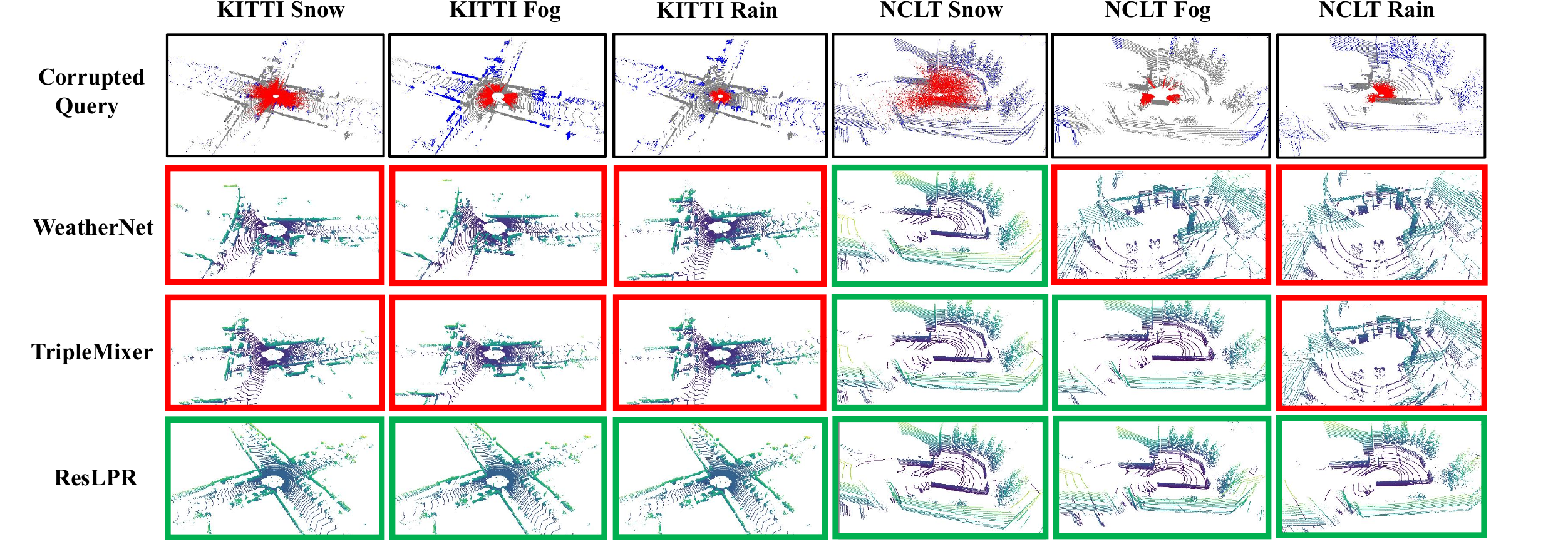}}
	 \caption{Qualitative visualizations of some corruption query examples, along with their top-1 retrieved matches on the WeatherKITTI and WeatherNCLT datasets using CVTNet. The \textcolor{red}{red points} in the point cloud signify the noise points, and the \textcolor{blue}{blue points} denote the lost points. \textcolor{red}{red boxes} indicate incorrect retrieval results, while \textcolor{green}{green boxes} denote correct retrievals. }
	 \label{query_vis}
\vspace{-0.5cm}
\end{figure*}

Besides evaluating preprocessing methods, our \ourben{} benchmark offers valuable analyses of existing LPR methods. For example, among the three tested LPR methods, we found that SC~\cite{SC} and LPSNet~\cite{LPSNet} demonstrated relatively better robustness on the WeatherKITTI dataset with less performance drop under corruption. In contrast, CVTNet~\cite{CVTNet} performed comparatively poorly. Even after applying \ours{}, the reported $\text{mSR}_{l}$ values for SC and LPSNet reach 0.949 and 0.915, while CVTNet, though significantly enhanced compared to other denoising methods and its corrupted result of 0.192, still achieves only 0.794. 
The performance gap originates from dense WeatherKITTI point clouds amplifying multi-view noise sensitivity in CVTNet, whereas SC and LPSNet maintain robustness via discrete processing and randomized sampling.
Additionally, all three LPR methods show more performance degradation under foggy conditions. This is likely due to fog particles obscuring many distant points, thereby reducing the effective range of the scene, which can also be seen from Fig.~\ref{query_vis}.

\subsubsection{Results on WeatherNCLT}
Table~\ref{ WeatherNCLT} reports the experimental results on the WeatherNCLT 32-beam LiDAR dataset in campus environments. As observed, all LPR methods exhibit relatively lower robustness under severe weather on WeatherNCLT compared to WeatherKITTI, due to the sparser LiDAR data and the extended temporal challenges in long-term LPR. Nevertheless, applying \ours{} consistently achieves the best restoration performance across all metrics for each LPR method under all tested conditions. Most notably, CVTNet fails on the corrupted data but achieves the best LPR performance after applying our \ours{}. 
These results show that using our restoration network, CVTNet's relative robustness ($\text{mSR}_{p}$) and absolute performance ($\text{Recall@N}$) significantly improve and outperform other LPR methods. This is attributed to CVTNet's designs for long-term place retrieval with sparser LiDAR data, as highlighted in its original work~\cite{CVTNet}. These findings also demonstrate that our restoration network plays a crucial role in enhancing the robustness of LPR methods under long-term seasonal weather changes.

\begin{table}[t]
\footnotesize
\captionsetup{aboveskip=2pt, belowskip=0pt}
\renewcommand\arraystretch{0.9}
\setlength{\tabcolsep}{3.9pt}
\centering
\caption{Ablation studies on each component of ResLPRNet}
\begin{tabular}{cccccccc}
\toprule
\multicolumn{2}{c}{\multirow{2}{*}{\textbf{Experiment}}} &\multirow{2}{*}{\textbf{WAT}} & \multirow{2}{*}{\textbf{CTG}} & \multicolumn{4}{c}{\textbf{Metric}} \\
\cmidrule(lr){5-8}
  & & & & \textbf{AUC↑} & \textbf{F1↑} & \textbf{R@1↑} & \textbf{R@5↑} \\
\midrule
\multirow{4}{*}{\text{CVTNet \cite{CVTNet}}} 
&\uppercase\expandafter{[A]} &  &  &0.310 &0.305 &0.598  &0.724 \\
&\uppercase\expandafter{[B]} &\checkmark &  &0.373 &0.345 &0.657 &0.776   \\
&\uppercase\expandafter{[C]} &  & \checkmark   &0.520 &0.478  &0.732  &0.826  \\
&\uppercase\expandafter{[D]} & \checkmark  &\checkmark   &\textbf{0.578} &\textbf{0.522} &\textbf{0.772}  &\textbf{0.856} \\
\bottomrule
\end{tabular}
\label{ablation_component}
\vspace{-0.5cm}
\end{table}

\subsubsection{Qualitative Analysis}

We further provide $\text{Recall@N}$ results on our benchmark in Fig.~\ref{recall@N}, showing that our \ours{} provides the most significant improvements for various LPR methods across diverse adverse weather. 
Additionally, Fig.~\ref{query_vis} visualizes the top-1 match results for each query scan processed by the denoising and restoration networks using CVTNet \cite{CVTNet}. The results show that \ours{} significantly outperforms the other two denoising methods in enhancing LPR robustness.

\subsection{Ablation Study}
In this section, we carry out an in-depth ablation study on the proposed ResLPRNet to explore the contributions of each component of our model and its restoration robustness under different severity grades.

\begin{figure}[t]
  \captionsetup{aboveskip=2pt, belowskip=0pt}
  \centerline{\includegraphics[width=\linewidth]{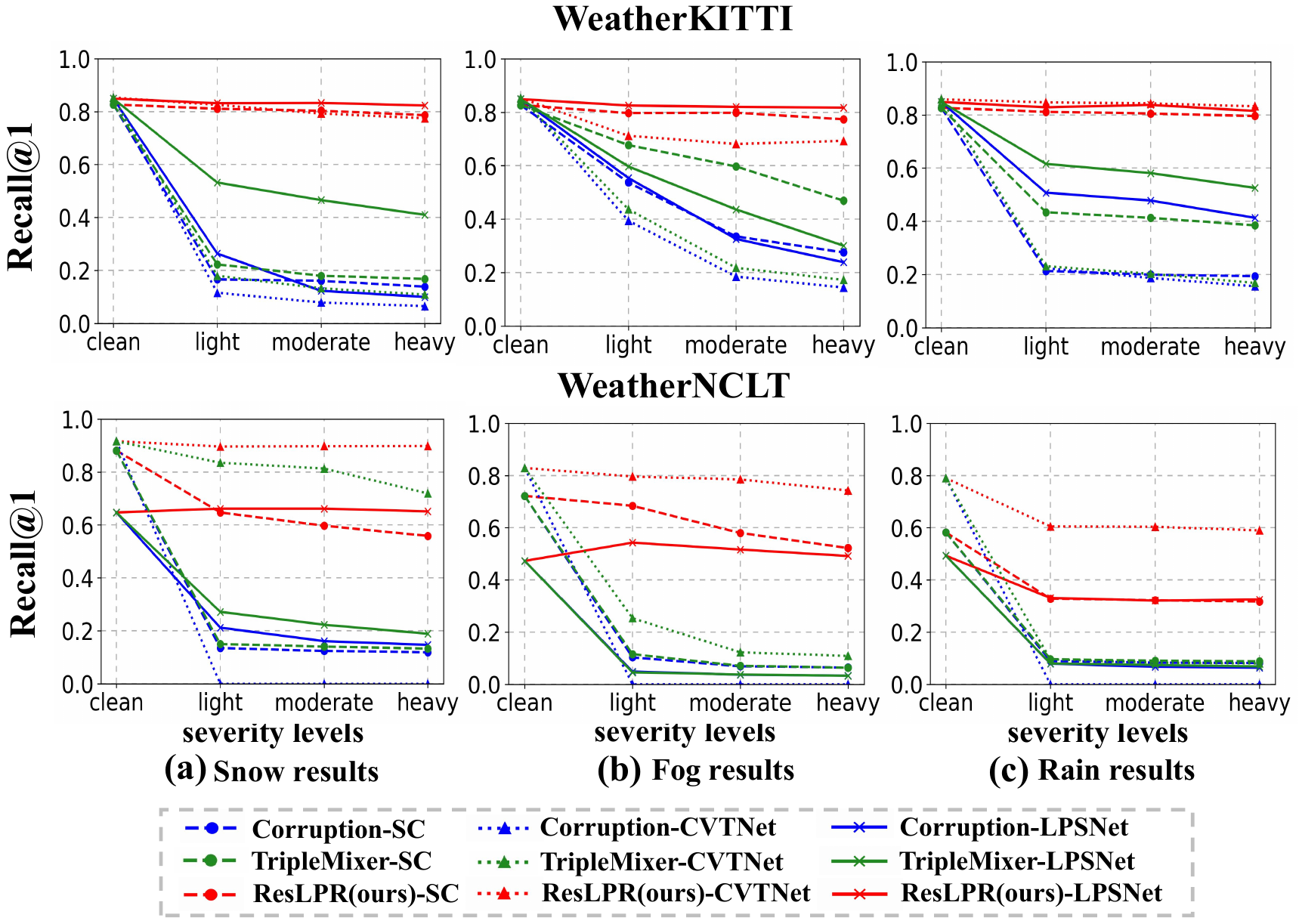}}
	\caption{LPR results under different types and severity levels of corruption. (a) Snow results (b) Fog results (c) Rain results}
	\label{ablation_level}
\vspace{-0.5cm}
\end{figure}

\subsubsection{Effects of Each Component}

We report the average results of CVTNet \cite{CVTNet} under moderate corruption in three different adverse weather conditions on the WeatherKITTI dataset. As presented in Table~\ref{ablation_component},Experiments [A], [B] and [C] represent the results of our network processing corrupted point clouds for place recognition without the WAT block and CTG block, or with one of them omitted individually. The comparison between the two sets of experiments involving the CTG module reveals its significant impact, with the $\text{AUC}$ increasing by nearly 0.2, and improvements observed in other metrics as well. The comparison between experiments [A] and [B], as well as between experiments [C] and [D], demonstrates that the WAT module also enhances performance across multiple metrics.
These results suggest that both the WAT and CTG blocks are critical to our \ours{}, with each contributing substantially to the model's restoration capability. 

\subsubsection{Effects of Weather-Induced Corruption Levels}

Ablation experiments across WeatherKITTI and WeatherNCLT with three corruption types at multiple severity levels. Fig.~\ref{ablation_level} reveal our method sustains high LPR performance under increasing severity, while other preprocessing method degrade, exhibiting superior generalization.

\subsection{Runtime}
We present the runtime of \ours{} to show its low computational cost and suitability as a plug-and-play module for LPR methods. Tests were conducted on an Intel i7-14700K CPU and Nvidia RTX 4070 GPU, using 1,200 LiDAR scans from our weather-induced LPR benchmark. The total parameter count of \ours{} is 31.65\,M, with an average runtime of 58\,ms, achieving real-time performance.

\section{Conclusion}
\label{sec:conclusion}

This paper introduces \ours{}, a point cloud restoration network designed to enhance LiDAR-based place recognition (LPR) models under adverse weather conditions. \ours{} exploits a WaveTransformer block that isolates weather-induced noise while preserving structural information and a ContextGuide block that learns point cloud patterns across various weather scenarios. Additionally, we present ResLPR, a novel public benchmark for evaluating LPR performance in adverse weather, providing the WeatherKITTI and WeatherNCLT datasets and a new protocol for assessing the effects of LiDAR data corruption on LPR methods. Experimental results demonstrate that \ours{} significantly improves LPR robustness across diverse weather conditions, outperforming other preprocessing methods under varying levels of weather-induced degradation in both short-term and long-term LPR tasks. We validate the effectiveness of key blocks within our model, which also achieves real-time performance. We hope that our restoration approach and benchmark can serve as a foundation for further advancements in LPR research under adverse weather conditions.

\bibliographystyle{ieeetr}

\bibliography{new}

\end{document}